\documentclass[graybox]{svmult}

\usepackage{titlesec}
\titleformat{\paragraph}
{\normalfont\normalsize\bfseries}{\theparagraph}{1em}{}
\titlespacing*{\paragraph}
{0pt}{3.25ex plus 1ex minus .2ex}{1.5ex plus .2ex}
\usepackage{mathptmx}       % selects Times Roman as basic font
\usepackage{helvet}         % selects Helvetica as sans-serif font
\usepackage{courier}        % selects Courier as typewriter font
\usepackage{type1cm}  
\usepackage{url}
\usepackage{amsmath}
\usepackage{hyperref}
\usepackage{csquotes}
\usepackage{graphicx}
\usepackage{array}
\usepackage{tabu}
\usepackage{float}
\usepackage{cite}
\usepackage{multirow,multicol}   % activate if the above 3 fonts are
\usepackage{makeidx}         % allows index generation
\usepackage{graphicx}        % standard LaTeX graphics tool
\usepackage{multicol}        % used for the two-column index
\usepackage[bottom]{footmisc}% places footnotes at page bottom

\makeindex             % used for the subject index
\begin{document}

\title*{A Study of Feature Extraction techniques for Sentiment Analysis}
\author{Avinash M and Sivasankar E*}
\institute{Avinash M \at National Institute of Technology,Tiruchirapalli , Tanjore Main Road, National Highway 67, Near BHEL Trichy, Tiruchirappalli, Tamil Nadu, 620015. \email{avinash.sai001@gmail.com}
\and Sivasankar E \at National Institute of Technology,Tiruchirapalli , Tanjore Main Road, National Highway 67, Near BHEL Trichy, Tiruchirappalli, Tamil Nadu, 620015. \email{sivasankar@nitt.edu}}
\maketitle

\abstract*{Each chapter should be preceded by an abstract (10--15 lines long) that summarizes the content. The abstract will appear \textit{online} at \url{www.SpringerLink.com} and be available with unrestricted access. This allows unregistered users to read the abstract as a teaser for the complete chapter. As a general rule the abstracts will not appear in the printed version of your book unless it is the style of your particular book or that of the series to which your book belongs.
Please use the 'starred' version of the new Springer \texttt{abstract} command for typesetting the text of the online abstracts (cf. source file of this chapter template \texttt{abstract}) and include them with the source files of your manuscript. Use the plain \texttt{abstract} command if the abstract is also to appear in the printed version of the book.}

\abstract{Sentiment Analysis refers to the study of systematically extracting the meaning of subjective text . When analysing sentiments from the subjective text using Machine Learning  techniques,feature extraction becomes a significant part. We perform a study on the performance of feature extraction techniques TF-IDF(Term Frequency-Inverse Document Frequency) and Doc2vec (Document to Vector) using Cornell movie review datasets, UCI sentiment labeled datasets, stanford movie review datasets,effectively classifying the text into positive and negative polarities by using various pre-processing methods like  eliminating Stop Words and Tokenization which increases the performance of sentiment analysis in terms of accuracy and time taken by the classifier.The features obtained after applying  feature extraction techniques on the text sentences are trained and tested using the classifiers Logistic Regression,Support Vector Machines,K-Nearest Neighbours , Decision Tree and Bernoulli Naïve Bayes.}

\section{Introduction}
\label{sec:1}
In Machine Learning,a feature refers to the information which can be extracted from any data sample.A feature uniquely describes the properties possessed by the data.The data used in machine learning consists of features projected onto a high dimensional feature space.These high dimensional features must be mapped onto a small number of low dimensional variables that preserves information about the data as much as possible.Feature extraction is one of the dimensionality reduction techniques used in machine learning to map higher dimensional data onto a set of low dimensional potential features.Extracting informative and essential features  greatly enhances the performance of machine learning models and reduces the computational complexity.

The growth of modern web applications like facebook,twitter persuaded users to express their opinions on products,persons and places. Millions of consumers review the products on online shopping websites like amazon,flipkart.These reviews act as valuable sources of information that can increase the quality of services provided.With the growth of enormous amount of user generated data,lot of efforts are made to analyze the sentiment from the consumer reviews.But analyzing unstructured form of data and extracting sentiment out of it requires a lot of natural language processing(nlp) and text mining methodologies.Sentiment analysis attempts to derive the polarities from the text data using nlp and text mining techniques.The classification algorithms in machine learning require the most appropriate set of features to  classify the text as positive polarity and negative polarity.Hence, feature extraction plays a prominent role in sentiment analysis.

\section{Literature Review}
\label{sec:2}
% Always give a unique label
% and use \ref{<label>} for cross-references
% and \cite{<label>} for bibliographic references
% use \sectionmark{}
% to alter or adjust the section heading in the running head
Bo Pang, Lillian Lee, and Shivakumar Vaithyanathan \cite{pang:lee:shivakumar} have conducted a study on sentiment analysis using machine learning techniques.They compared the performance of machine learning techniques with human generated baselines and proposed that machine learning techniques are quite good in comparison to human generated baselines.A. P. Jain and V. D. Katkar \cite{jain:katkar} have performed sentiment analysis on twitter data using data mining techniques.They analyzed the performance of various data mining techniques and proposed that data mining classifiers can be a good choice for sentiment prediction.Tim O'Keefe and Irena Koprinska \cite{keefe:irena} have done research on feature selection and weighting methods in sentiment Analysis. In their research they have combined various feature selection techniques with feature weighing methods to estimate the performance of classification algorithms. Shereen Albitar, Sebastien Fournier, Bernard Espinasse \cite{Albitar:Fournier:Espinasse} have a proposed an effective TF-IDF based text-to-text semantic similarity measure for text classification. Quoc Le and Tomas Mikolov \cite{quoc:mikolov} introduced distributed representation of sentences and documents  for text classification.
Parinya Sanguansat\cite{Parinya} performed paragraph2vec based sentiment analysis on social media for business in Thailand.Izzet Fatih Şenturk and Metin Bilgin \cite{Metin:İzzet} performed sentiment analysis on twitter data using doc2vec. 
Analysing all the above techniques  gave us deep insight about various methodologies used in sentiment analysis.Therefore, we propose a comparison based study on the performance of feature techniques used in sentiment analysis and the results are described in the following sections.

The  proposed study involves two different features extraction techniques TF-IDF and Doc2Vec. These techniques are used to extract features from the text data. These features are then used in identifying the polarity of text data using classification algorithms.The performance evaluation of TF-IDF and Doc2vec are experimented on 5 benchmark datasets of varying sizes. These datasets contains user reviews from various domains. Experimental results shows that the performance of feature extraction techniques trained and tested using these benchmark datasets.A lot of factors affects the sentiment analysis of text data.When the text data is similar it becomes difficult to extract suitable features.If the features extracted are not informative,it significantly affects the performance of classification algorithms.When the patterns lie closest to each other,drawing
separating surfaces can be tricky for the classifiers.Classifiers like SVM aim to
maximize the margin between two classes of interest.
\section{Methodology}
\begin{figure}[!t]
	% For example, with the graphicx style use
	\sidecaption
	\includegraphics[width=75mm,scale=0.5]{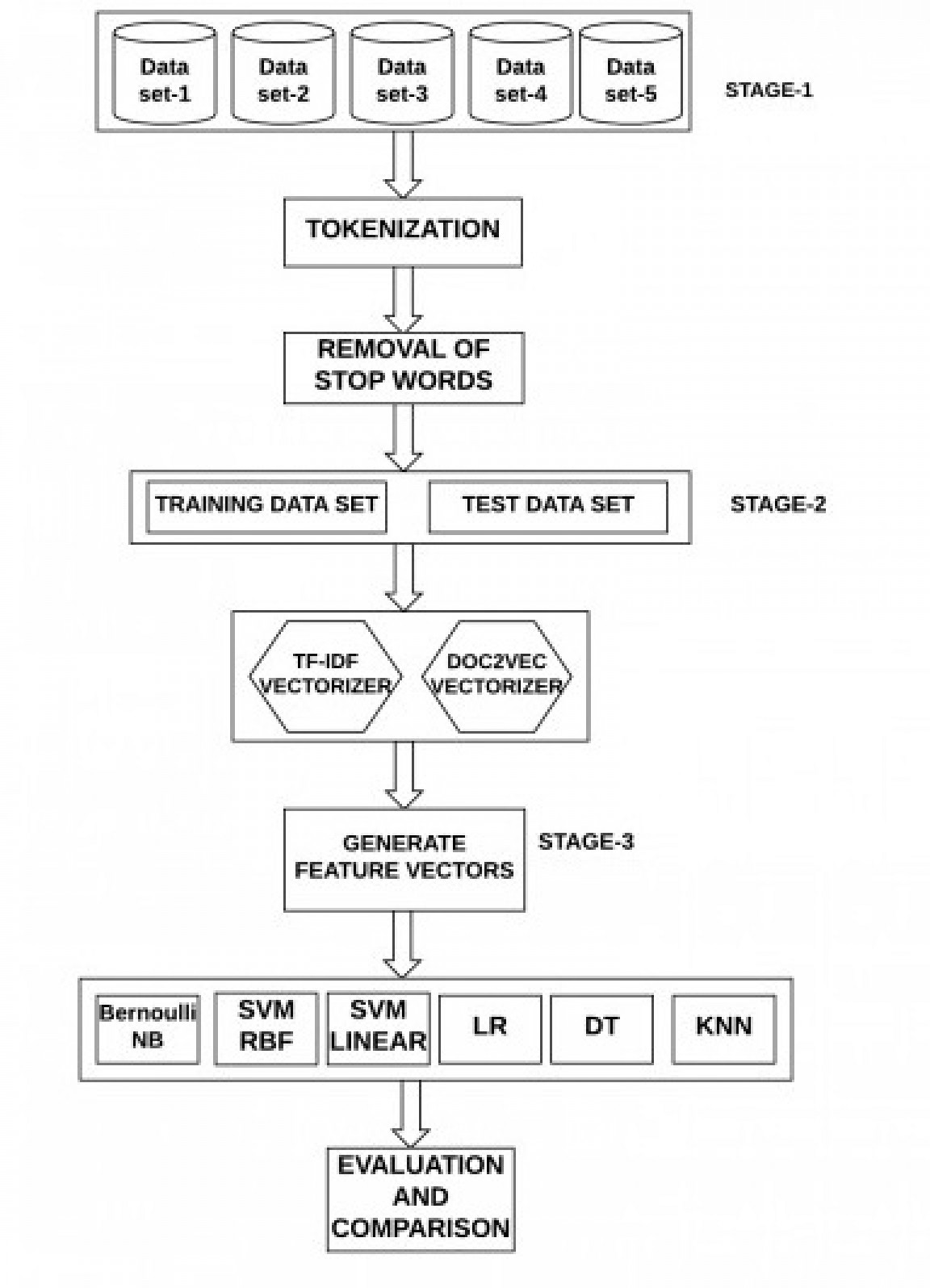}
	%
	% If no graphics program available, insert a blank space i.e. use
	%\picplace{5cm}{2cm} % Give the correct figure height and width in cm
	%
	\caption{Experimental Setup}
	\label{fig:1}       % Give a unique label
\end{figure}

The  performance evaluation of feature extraction techniques TF-IDF and Doc2Vec  requires datasets of various sizes. The tests are conducted on the publicly available datasets related to movie reviews and reviews related to other commercial products. The performance is evaluated by varying the sizes of datasets and the accuracy of each of the two feature extraction techniques is measured using the classifiers.The first data set is a part of  movie review dataset \cite{Maas:Andrew:Christopher} consisting of 1500 reviews.The second dataset used is Sentiment Labelled dataset\cite{Dimitrios:Misha:Freitas:Padhraic} consisting of  2000 reviews taken from UCI repository of datasets.The third dataset used is a part of polarity dataset v2.0\cite{pang:Lillian} taken from CS cornell.edu.The fourth dataset sentence polarity dataset v1.0\cite{pang:lee} taken from CS cornell.edu.The fifth dataset used is Large movie review dataset\cite{Maas:Andrew:Christopher}  taken from ai.stanford.edu. The reviews are categorized into positive and negative. We perform analysis on reviews in 3 stages – Preprocessing,Feature Extraction and classification.Figure \ref{fig:1} shows the experimental setup for sentiment analysis. 
\subsection{Pre-processing}
We perform Pre-processing in two stages- Tokenization and by eliminating stop words.
Tokenization refers to the process of breaking the text into meaningful data that retains the information about the text.Text data contain some unwanted words that does not give any meaning to the data called as stop words.These words are removed from the text during feature extraction procedures.
\subsection{Feature Extraction}
\begin{figure}[!t]
		% For example, with the graphicx style use
		\sidecaption[!t]
		\includegraphics[width=75mm,scale=0.2]{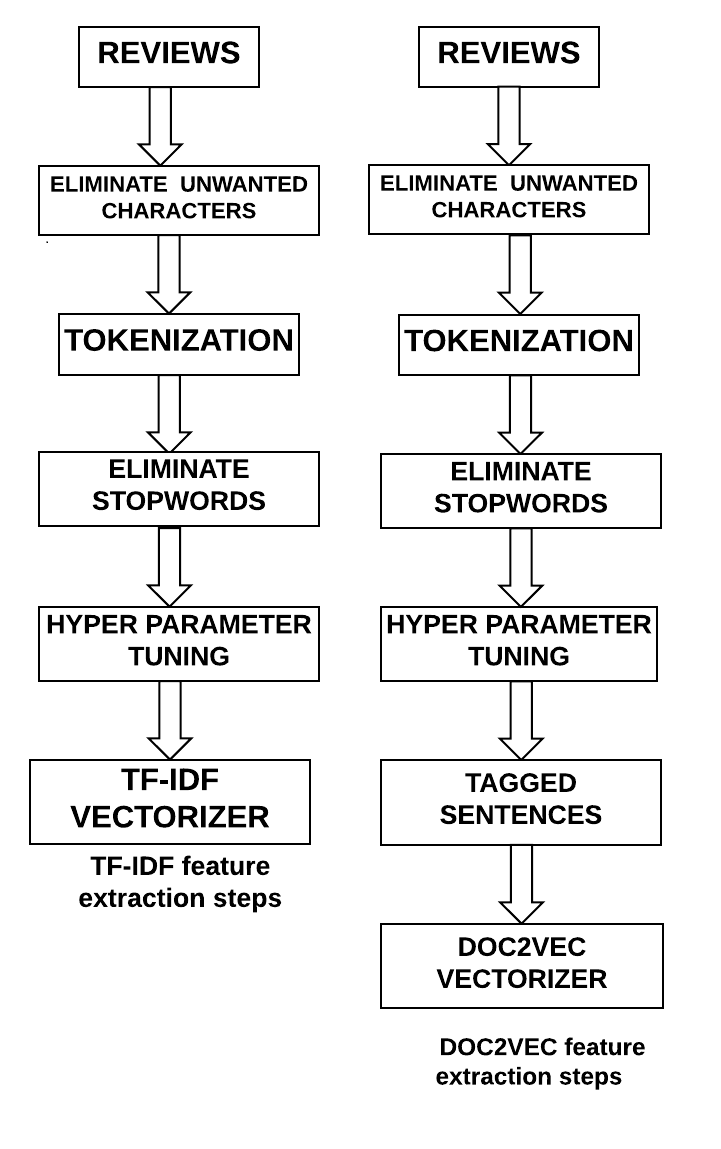}
		%
		% If no graphics program available, insert a blank space i.e. use
		%\picplace{5cm}{2cm} % Give the correct figure height and width in cm
		%
		\caption{Feature Extraction Procedure}
		\label{fig:2}       % Give a unique label
	\end{figure}

Two techniques are used for feature extraction -TF-IDF and Doc2vec.Figure \ref{fig:2} shows the steps involved in extracting features using both the techniques. 
\subsubsection{TF-IDF} 
TF-IDF is short form for term frequency-inverse document frequency.TF-IDF is one of the largely used methods in information retrieval and text mining.TF-IDF is a weight metric which determines the importance of word for that document.

\paragraph{TF}
Term Frequency measures number of times a particular term t occured in a document d.
Frequency increases when the term has occured multiple times.TF is calculated by taking ratio of frequency of term t in document d to number of terms in that particular document d.
\begin{equation}
 TF(t,d)=\frac{\text{Number of times term $t$ appears in a document $d$}}{\text{Total number terms in a document $d$}}\
\end{equation}
\paragraph{IDF}
TF measures only the frequency of a term t.Some terms like stop words occur multiple times but may not be useful.Hence Inverse Document Frequency(IDF) is used to measure term's importance.IDF gives more importance to the rarely occuring terms in the document d. IDF is calculated as:
\begin{equation}
IDF(t) = log_e \frac{\text{Total number of documents }}{\text{Total number of documents with term $t$ in it}}
\end{equation}
The final weight for a term t in a document d is calculated as: 
	\begin{equation}
	TF-IDF(t,d) =  TF(t,d) \times IDF(t)
	\end{equation}

\subsubsection{Doc2Vec}
An extended version to word2vec, doc2vec model was put forward by Le and Miklov(2014)\cite{quoc:mikolov} to improve the learning of embeddings from word to word sequences. doc2vec can be applied for word n-gram, sentence, paragraph or document. Doc2vec is a set of approaches to represent documents as fixed length low  dimensional  vectors. Doc2Vec is a three layer neural network with an input ,one hidden layer and an ouput layer.Doc2vec was proposed in two forms: dbow and dmpv.In Word2vec two algorithms continuous bag of words (CBOW) and skip-gram(SG) algorithms are implemented using deep Learning, in Doc2vec these algorithms correspond to distributed memory(DM) and distributed bag of words(DBoW).

\subsection{Classification}
Classifiers are trained on training data sets with the features obtained from the feature extraction techniques TF-IDF and Doc2vec and with the corresponding output labels from the dataset. The test data is tested with the trained classifiers to predict the sentiments and the accuracy is measured for test data.Classifiers Logistic Regression , KNearest Neighbors, Decision Tree, Bernoulli Naïve Bayes and support vector machines with linear and rbf kernels are used for sentiment analysis of the datasets.
\subsubsection{Logistic Regression(LR)}
In linear regression we try to predict the value $y^\textsuperscript{(i)}$ for the $i^\textsuperscript{th}$ training sample using a linear function $Y = h_{\theta}(x) = {\theta}^{T} x$. This  is clearly not a great solution to predict binary-valued labels.$(y^{(i)} \in \{0,1\} )$.Logistic regression use a different hypothesis  that predicts the probability that a given example belongs to class \enquote{1} and the probability that the example belongs to class \enquote{0} . Logistic Regression is formulated as:
\begin{eqnarray}
 P(y=1|x) = \frac{1}{1+e^{-{\theta}^{T} x}}\\
 P(y=0|x) = 1-\ h_{\theta}(x)
\end{eqnarray}
\subsubsection{K-Nearest Neighbors(KNN)}
In KNN,the input is represented in the feature space by k-closest training samples. KNN uses the procedure of class membership. Each object is classified by taking its neighbors votes and the object is assigned to class which has got maximum number of votes.The training examples are represented as vectors in a multi-dimensional feature space with a class label. The training of the KNN algorithm involves storing training samples  as feature vectors with their corresponding labels. k is decided by the user during classification and the unlabeled data is assigned to class which is more frequent among k of the training samples nearest to that unlabeled sample.Euclidean distance is used to calculate the distance.
\subsubsection{Naive Bayes(NB)}
Bayes theorem forms the basis for Naive Bayes classifier.It assumes that features are strongly independent and calculates the probability.Multi variate Bernoulli event model is one of the naive bayes classifiers used in sentimental analysis. If $X_{i}$ is a boolean  expressing the occurrence or absence of the  i’th term from the vocabulary , then the likelihood of a document given a class $C_{k}$ is given by:
\begin{equation}
 P(x | C_{k} ) = \prod_{i=1}^n p_{k_i}^{x_i}  (1-p_{k_i})^{1-x_{i}}
\end{equation}
\subsubsection{Support Vector Machines(SVM)}
The training samples are represented as points in the feature space.SVM performs classification by separating the points with a set of margin planes.The boundary hyperplane is chosen which maximizes the distance to the training samples.Support vectors are the points that determine the margin planes. Data which can be separated linearly is classified using Linear kernel and the data which is not linearly separable is classified using RBF kernel. 
\subsubsection{Decision Tree(DT)}
Decision Trees performs classification by using yes or no conditions.A decision tree consists of edges and nodes.The root node  doesn't contain incoming edges,nodes which contain out going edges are called test nodes,nodes with no outgoing nodes are called decision nodes.Edges represent the conditions.Each decision node holds a class label.When the unlabeled data samples are to be classfied,they pass through series of test nodes finally leading to the decision node with a class label and the class label is assigned to that unlabeled data sample.

\section{Experimental Results}
The  performance of TF-IDF and Doc2Vec is verified by experimenting the feature extraction techniques on 5 different datasets of varying sizes.
First the experiment is done on a small data set(700 positive and 800 negative reviews) of movie reviews taken from Large polarity dataset 1.0\cite{Maas:Andrew:Christopher}. Secondly, the experiment is done on Sentiment Labelled dataset\cite{Dimitrios:Misha:Freitas:Padhraic}(2000 reviews) of UCI data repository. The third experiment is done on a corpus (100 negative and 100 positive documents) made from the  polarity dataset v2.0\cite{pang:Lillian}(1000 negative and 1000 positive documents) taken from the CS.cornell.edu. The fourth experiment is done on sentence polarity dataset v1.0(5331 positive and 5331 negative reviews) taken from CS.Cornell.edu\cite{pang:lee} . Fifth experiment is done on  Large movie review dataset \cite{Maas:Andrew:Christopher}(25000 positive reviews and 25000 negative reviews) taken from the  ai.stanford.edu. Using  regular expressions and string processing methods sentences and labels are separated from the datasets. The sentences obtained are fed into feature extraction techniques TF-IDF  and Doc2Vec to generate vector(real numbers) features for each sentence.The Split of training and testing samples is done by either hold out method where 50\% data is used for training and 50\% data is used for testing or by 10-fold cross validation(CV) where 9 folds are used for training and 1 fold is used for testing.Table \ref{tab:1} shows the methods used for splitting training and test datasets. 

% Use the \index{} command to code your index words
\vspace{-1em}
\begin{table}
\caption{Methods used for training and testing}
\label{tab:1}       % Give a unique label
%
% Follow this input for your own table layout
%
\begin{tabular}{p{3cm}p{1.7cm}p{1.7cm}p{1.7cm}p{1.7cm}p{1.7cm}}
\hline\noalign{\smallskip}
Datasets & Dataset-1 & Dataset-2 & Dataset-3 & Dataset-4 & Dataset-5  \\
\noalign{\smallskip}\svhline\noalign{\smallskip}
Method used & Hold out & 10 fold CV & 10 fold CV  & 10 fold CV  &  Hold out \\
Training samples & 800 & 9 folds & 9 folds  & 9 folds  & 25000 \\
Testing samples & 700 & 1 fold & 1 fold & 1 fold & 25000 \\
\noalign{\smallskip}\hline\noalign{\smallskip}
\end{tabular}
\end{table}
\vspace{-3em}
\begin{table}
	\caption{Hyper parameters tuned for TF-IDF Vectorizer}
	\label{tab:2}       % Give a unique label
	%
	% Follow this input for your own table layout
	%
	\begin{tabular}{p{3cm}p{1.7cm}p{1.7cm}p{1.7cm}p{1.7cm}p{1.7cm}}
		\hline\noalign{\smallskip}
		Parameters and Datasets & Dataset-1 & Dataset-2 & Dataset-3 & Dataset-4 & Dataset-5  \\
		\noalign{\smallskip}\svhline\noalign{\smallskip}
		min\_df & 5 & 5 & 5  & 5  &  5 \\
		max\_df & 0.8 * n\_d & 0.8 * n\_d & 0.8 * n\_d  & 0.8 * n\_d  & 0.8 * n\_d\\
		encoding & utf-8 & utf-8 & utf-8 & utf-8 & utf-8 \\
		sublinear\_df & True  & True & True & True  & True  \\
		use\_idf &True &True & True&True &True \\
		stop words & English & English & English& English &English \\
		\noalign{\smallskip}\hline\noalign{\smallskip}
	\end{tabular}
\end{table}
\vspace{-3em}
\begin{table}
	\caption{Hyper parameters tuned for Doc2Vec Vectorizer}
	\label{tab:3} 
\begin{tabular}{p{3cm}p{1.7cm}p{1.7cm}p{1.7cm}p{1.7cm}p{1.7cm}}
	\hline\noalign{\smallskip}
	Parameters and Datasets & Dataset-1 & Dataset-2 & Dataset-3 & Dataset-4 & Dataset-5  \\
	\noalign{\smallskip}\svhline\noalign{\smallskip}
	min count & 1 & 1 & 1  & 1  &  1 \\
	window size & 10 & 10 & 10  & 10  & 10  \\
	vector size & 100 & 100 & 100 & 100 & 100 \\
	sample & 1e-4  & 1e-5 & 1e-5 & 1e-5  & 1e-4\\
	negative & 5 & 0 & 5 & 5 & 5\\
	workers & 7 & 1 & 1 & 1 & 7 \\
	dm & 1 & 0 & 0 & 0 & 1 \\
	\noalign{\smallskip}\hline\noalign{\smallskip}
\end{tabular}
\end{table}
Each training and testing data set obtained by any of the above methods are fed into feature extraction techniques TF-IDF and Doc2Vec to generate vectors.The TF-IDF vectorizer used is tuned with hyper parameters min\_df (min number of times term t has to occur in all sentences),n\_d = Number of sentences in a training or testing
corpus, max\_df (maximum number of times a term t can occur in all the sentences  which is calculated as max\_df * n\_d),
Encoding(encoding used), sublinear\_tf (use of term frequency) , use\_idf (use of inverse document frequency), stopwords(to eliminate stopwords). The Doc2Vec vectorizer used is tuned with hyper parameters min\_count(minimum number of times a word has to occur),window size(maximum distance between predicted word and context words used for prediction),vector size(size of vector for each sentence),sample( threshold for configuring which higher  frequency words are randomly downsampled),negative(negative sampling is used for drawing noisy words),workers(number of workers used to extract feature vectors),dm(type of training algorithm used - 0 for distributed bag of words and 1 for distributed model), Epochs(number of iterations for training).Tables \ref{tab:2} and \ref{tab:3} shows hyper paramaters tuned for both the techniques.
\section{Performance Analysis}
To estimate the performance of TF-IDF and Doc2Vec on classification algorithms we use accuracy measure as the metric.
Accuracy is calculated as  the ratio of number of reviews that are correctly predicted to the total number of reviews. Let us assume 2 $\times$ 2  matrix as defined in Table \ref{tab:4}. Tables \ref{tab:5} and \ref{tab:6} shows the accuracy of all classifiers for each feature extraction technique trained and tested on all the datasets. 
\vspace{-1em}
\begin{table}
	\caption{Contingency matrix for sentiment analysis}
	\label{tab:4}       % Give a unique label
	%
	% Follow this input for your own table layout
	%
	\begin{tabular}{p{3cm} p{3cm} p{3cm} p{3cm}}
		\hline\noalign{\smallskip}
		& & &\textbf{Predicted sentiment}   \\
		\noalign{\smallskip}\svhline\noalign{\smallskip}
	    &   & +ve sentiment  & -ve sentiment \\
	    \noalign{\smallskip}\hline\noalign{\smallskip}
	    &+ve sentiment  & True Positive(TP)	 & False Negative(FN)	 \\
	    \textbf{Actual Sentiment}	& -ve sentiment 	& False Positive(FP) & True Negative(TN) \\
		\noalign{\smallskip}\hline\noalign{\smallskip}
	\end{tabular}
\end{table}
\vspace{-3em}
\begin{table}[!h]
	\caption{\textbf{Accuracy of classifiers}}
	\label{tab:5} 
	\begin{tabular}{p{3cm}p{1.5cm}p{1.5cm}p{1.5cm}p{1.5cm}p{1.5cm}p{1.5cm}}
		\hline
		Datasets & Dataset-1 & Dataset-1 &Dataset-2 & Dataset-2 & Dataset-3 &Dataset-3\\
		\hline
		Classifiers &TF-IDF & Doc2Vec & TF-IDF & Doc2Vec  & TF-IDF  &  Doc2Vec \\
		\hline
		logistic regression & \textbf{83.428} & 86.1428 &  81.950  & \textbf{77.400}  & 98.864 & \textbf{98.864} \\
		SVM-rbf & 75.142 & \textbf{86.85714} & \textbf{82.350} &\textbf{77.400} & \textbf{98.887} & \textbf{98.864} \\
		SVM-linear &78.000  & 85.71428 & 77.400& \textbf{77.400}  & 98.864 & \textbf{98.864} \\
		KNN(n=3) & 64.714 & 74.142 & 77.400 & 70.05 & 98.793 &98.816 \\
		DT & 65.571 & 70.285 & 76.950 & 64.850 & 97.633 & 97.065 \\
		bernoulli-NB & 80.714 & 83.71428 & 81.050 & 76.850 & 98.840 & 97.823 \\
		\hline
	\end{tabular}
\end{table}
\vspace{-1em}
\begin{table}[!h]
	\caption{\textbf{Accuracy of classifiers}}
	\label{tab:6} 
	\begin{tabular}{p{3cm}p{1.5cm}p{1.5cm}p{1.5cm}p{1.5cm}}
		\hline
		Datasets & Dataset-4 & Dataset-4 &Dataset-5 & Dataset-5 \\
		\hline
		Classifiers &TF-IDF & Doc2Vec & TF-IDF & Doc2Vec  \\
		\hline
		logistic regression &82.633  &  \textbf{99.98124} &  \textbf{88.460}  &86.487  \\
		SVM-rbf &\textbf{82.820}  & \textbf{99.98124} & 67.180 & \textbf{86.760} \\
		SVM-linear & 82.596 & \textbf{99.98124} & 87.716 & 86.472 \\
		KNN(n=5) & 82.417 & \textbf{99.98124} & 68.888 & 74.592 \\
		DT & 74.137 & 99.97186& 71.896 &67.140  \\
		bernoulli-NB & 79.330 & \textbf{99.98124} & 82.020 & 80.279 \\
		\hline
	\end{tabular}
\end{table}
\begin{figure}[!t]
	% For example, with the graphicx style use
	\sidecaption[!t]
	\includegraphics[width=75mm,scale=0.5]{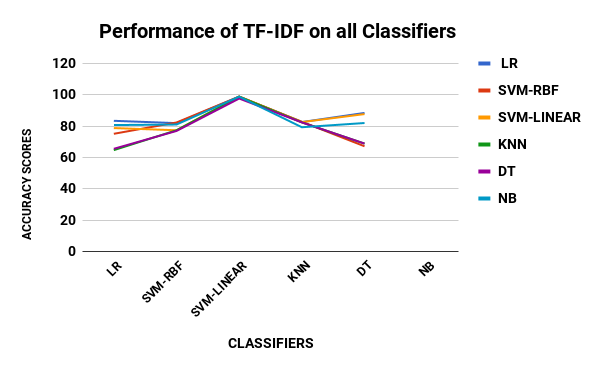}
	%
	% If no graphics program available, insert a blank space i.e. use
	%\picplace{5cm}{2cm} % Give the correct figure height and width in cm
	%
	\caption{TF-IDF performance}
	\label{fig:3}       % Give a unique label
\end{figure}
\begin{figure}[!t]
	% For example, with the graphicx style use
	\sidecaption[!t]
	\includegraphics[width=75mm,scale=0.5]{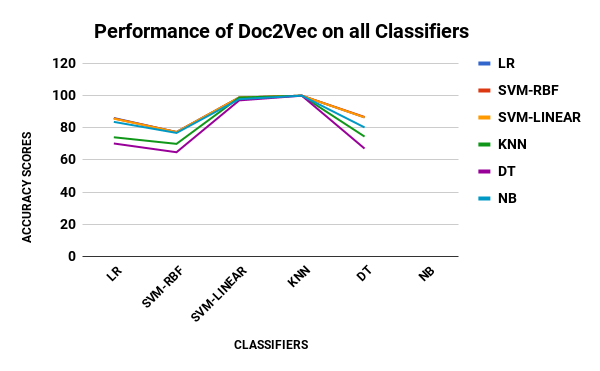}
	%
	% If no graphics program available, insert a blank space i.e. use
	%\picplace{5cm}{2cm} % Give the correct figure height and width in cm
	%
	\caption{Doc2Vec performance}
	\label{fig:4}       % Give a unique label
\end{figure}

Figure \ref{fig:3} shows the plot of accuracy for all classifiers using TF-IDF. Figure \ref{fig:4} shows the plot of accuracy for all classifiers using Doc2Vec.
Based on the accuracy values from Tables \ref{tab:5} and \ref{tab:6},
\begin{enumerate}
	\item
	For the first dataset, Doc2Vec performed better than TF-IDF for all classifiers.The accuracy is highest for Logistic Regression in case of TF-IDF whereas SVM with rbf kernel achieved highest accuracy in case of Doc2Vec.
	\item
	For the second dataset, TF-IDF performed better than Doc2Vec.The accuracy is highest for SVM with rbf kernel in case of TF-IDF whereas LR,SVM with rbf and linear kernels achieved highest accuracy in case of Doc2Vec. 
	\item
	For the third dataset, TF-IDF and Doc2Vec achieved similar accuracies.SVM with rbf kernel achieved highest accuracy in case of TF-IDF whereas LR,SVM with rbf and linear kernels achieved highest accuracy in case of Doc2Vec.
	\item
	For the fourth dataset, Doc2Vec performed better than TF-IDF for all classifiers.The accuracy is highest for SVM with rbf kernel in case of TF-IDF whereas LR,SVM with rbf,linear kernels,Bernoulli NB achieved highest accuracy in case of Doc2Vec.
	\item
	For the fifth dataset, TF-IDF performed slightly better than Doc2Vec.The accuracy is highest for LR in case of TF-IDF whereas SVM with rbf kernel achieved highest accuracy for Doc2Vec. 
\end{enumerate}
Considering  the performance analysis of all the classifiers on all datasets , we conclude that Logistic regression, SVM with linear and rbf  kernels performs better than all the other classifiers.
\section{Conclusion:}
The  purpose of this robust analysis is to provide deeper insight about the performance of feature extraction techniques TF-IDF and Doc2Vec . Based on the accuracy measures for all the datasets, Doc2vec and TF-IDF  has achieved satisfactory performance for most of the datasets. But the accuracy scores of Doc2vec are better when compared to TF-IDF on most of the data sets. This work can be extended to test these techniques on unstructured data to analyze the performance of features extracted from both the methods.


\begin{thebibliography}{6}
	%
	\bibitem {pang:lee:shivakumar}
	B.Pang,L.Lee,S.Vaithyanathan: Thumbs up?: sentiment classification using machine learning techniques,In: Proceedings of the ACL-02 Conference on Empirical Methods in Natural Language Processing,Vol 10,Series. EMNLP '02, pp. 79-86, (2002).
	\bibitem{jain:katkar}
	A. P. Jain and V. D. Katkar : Sentiments analysis of Twitter data using data mining,
	In : International Conference on Information Processing (ICIP), pp. 807-810, (2015).
	\bibitem {keefe:irena}
    Irena Koprinska and	Tim O'Keefe: Feature selection and weighting methods in sentiment analysis. In: Proceedings of the 14th Australasian Document Computing Symposium, pp. 67-74, Sydney, Australia, (2009).
	\bibitem{Albitar:Fournier:Espinasse}
	Albitar S.,Espinasse B, Fournier S.: An Effective TF/IDF-Based Text-to-Text Semantic Similarity Measure for Text Classification.In: Benatallah B., Bestavros A., Manolopoulos Y,Vakali A,Zhang Y.(eds): Web Information Systems Engineering– WISE 2014. Lecture Notes in Computer Science, Vol 8786. Springer, Cham,(2014).
	\bibitem{quoc:mikolov}
	Quoc V. Le, Tomas Mikolov : Distributed Representations of Sentences and Documents, In: CoRR, Vol. abs/1405.4053, (2014).
	\bibitem{Parinya}
	Parinya Sanguansat: Paragraph2Vec-Based Sentiment Analysis on Social Media for Business in Thailand. In: 8th International Conference on Knowledge and Smart Technology (KST), pp. 175-178, (2016).
	\url{doi:10.1109/KST.2016.7440526}
	\bibitem{Metin:İzzet}
	M. Bilgin and I.F.Senturk: Sentiment analysis on Twitter data with semi-supervised Doc2Vec. In:International Conference on Computer Science and Engineering(UBMK), pp. 661-666, (2017). 
	\url{doi:10.1109/UBMK.2017.8093492}
	\bibitem {Maas:Andrew:Christopher}
	Andrew L. Maas, Andrew Y. Ng,Christopher Potts,Dan Huang,Peter T. Pham,Raymond E. Daly : Learning Word Vectors for Sentiment Analysis. In: Proceedings of the 49th Annual Meeting of the Association for Computational Linguistics: Human Language Technologies, Vol. 1, Series. HLT '11, pp. 142-150, Portland, Oregon ,(2011).
	\bibitem {Dimitrios:Misha:Freitas:Padhraic}
	Dimitrios Kotzias,Misha Denil, Nando de Freitas,Padhraic Smyth : From Group to Individual Labels Using Deep Features, In : Proceedings of the 21th ACM SIGKDD International Conference on Knowledge Discovery and Data Mining, Series. KDD '15, pp. 597-606, Sydney, NSW, Australia ,(2015).
    \bibitem{pang:Lillian}
    Bo Pang and Lillian Lee: A Sentimental Education: Sentiment Analysis Using Subjectivity Summarization Based on Minimum Cuts. In: Proceedings of the 42nd Annual Meeting on Association for Computational Linguistics, Series. ACL '04, articleno. 271, (2004).
	\bibitem{pang:lee}
    Bo Pang,Lillian Lee : Seeing Stars: Exploiting Class Relationships for Sentiment Categorization with Respect to Rating Scales.In :Proceedings of the 43rd Annual Meeting on Association for Computational Linguistics, Series. ACL'05, pp. 115-124, (2005).
\end{thebibliography}
\end{document}